\documentclass[journal]{IEEEtran}
\usepackage{graphicx}
\usepackage{subfigure}
\usepackage{url}
\usepackage{multirow}
\usepackage{mwe}
\usepackage{xcolor}
\usepackage{makecell}

\newcommand{\specialcell}[2][c]{%
  \begin{tabular}[#1]{@{}c@{}}#2\end{tabular}}
  
\ifCLASSINFOpdf
\else
\fi
\begin{document}
%
\title{Learning Normalized Inputs for Iterative Estimation in Medical Image Segmentation}
%
%
%

\author{Michal Drozdzal, Gabriel Chartrand, Eugene Vorontsov, Lisa Di Jorio, \\ An Tang, Adriana Romero, Yoshua Bengio, Chris Pal, Samuel Kadoury
\thanks{M. Drozdzal was with \'{E}cole Polytechnique de Montr\'{e}al, Montr\'{e}al, Montreal Institute for Learning Algorithms and Imagia Inc., Montr\'{e}al, e-mail:michal.drozdzal@umontreal.ca}
\thanks{G. Chartrand was with Universit\'{e} de Montr\'{e}al, Montr\'{e}al and Imagia Inc., Montr\'{e}al}
\thanks{L. Di Jorio was with Imagia Inc., Montr\'{e}al}
\thanks{A. Tang was with CHUM Research Center, Montr\'{e}al}
\thanks{E. Vorontsov and C. Pal were with \'{E}cole Polytechnique de Montr\'{e}al, Montr\'{e}al and Montreal Institute for Learning Algorithms}
\thanks{A. Romero and Y. Bengio were with Montreal Institute for Learning Algorithms}
\thanks{S. Kadoury was with \'{E}cole Polytechnique de Montr\'{e}al, Montr\'{e}al and CHUM Research Center, Montr\'{e}al}
}

%
%

\markboth{}%
{Shell \MakeLowercase{\textit{et al.}}: Bare Demo of IEEEtran.cls for IEEE Journals}
%



\maketitle

\begin{abstract}
In this paper, we introduce a simple, yet powerful pipeline for medical image segmentation that combines Fully Convolutional Networks (FCNs) with Fully Convolutional Residual Networks (FC-ResNets). We propose and examine a design that takes particular advantage of recent advances in the understanding of both Convolutional Neural Networks as well as ResNets. Our approach focuses upon the importance of a trainable pre-processing when using FC-ResNets and we show that a low-capacity FCN model can serve as a pre-processor to normalize medical input data. In our image segmentation pipeline, we use FCNs to obtain normalized images, which are then iteratively refined by means of a FC-ResNet to generate a segmentation prediction. As in other fully convolutional approaches, our pipeline can be used off-the-shelf on different image modalities. We show that using this pipeline, we exhibit state-of-the-art performance on the challenging Electron Microscopy benchmark, when compared to other 2D methods. We improve segmentation results on CT images of liver lesions, when contrasting with standard FCN methods. Moreover, when applying our 2D pipeline on a challenging 3D MRI prostate segmentation challenge we reach results that are competitive even when compared to 3D methods. The obtained results illustrate the strong potential and versatility of the pipeline by achieving highly accurate results on multi-modality images from different anatomical regions and organs.
\end{abstract}

\begin{IEEEkeywords}
Image Segmentation, Fully Convolutionl Networks, ResNets, Computed Tomography, Electron Microscopy, Magnetic Resonance Imaging.
\end{IEEEkeywords}

%
\IEEEpeerreviewmaketitle

\section{Introduction}
\label{sec:Intro}
\IEEEPARstart{S}{egmentation} is an active area of research in medical image analysis. With the introduction of Convolutional Neural Networks (CNNs), significant improvements in performance have been achieved in many standard datasets. For example, for the EM ISBI 2012 dataset~\cite{EM_data}, PROMISE12 challenge or MS lesions~\cite{MsLesions}, the top entries are built on CNNs \cite{RonnebergerFB15,0011QCH16,HavaeiDWBCBPJL15,Yu17}.

The common view on CNN models is based on a \textit{representation learning} perspective \cite{Greff16,Goodfellow-et-al-2016-Book}. This view assumes that a CNN is built from layers (convolutional operations and non-linearities) that learn increasing levels of abstraction. In \cite{ZeilerF13}, these different levels of abstraction were visualized, showing that in the first layer the network learns simple edge and blob detectors (e.g. Gabor-like filters), the second layer learns combination of these simple features, while the deeper layers learn to represent more complex object contours, such as faces or flowers. Moreover, in \cite{VeitWB16}, it was shown that removing any single layer of the network after training/finetuning can significantly harm the network's performance, implying that the transformations learnt by a CNN layer are very different from an identity mapping.

Recently, a new class of models called Residual Networks (ResNets) have been introduced \cite{HeZRS15,HeZR016}. ResNets are built from hundreds of residual blocks. Each residual block is composed of two paths: the first one applies a series of nonlinear transformations (typically two or three transformations composed of Batch Normalization \cite{SzegedyLJSRAEVR14}, convolution and a ReLu non-linearity), while the second one is an identity mapping. These two paths are summed up at the end of a residual block. This small architectural modification has three important implications. First, the gradient can flow uninterrupted allowing parameters to be updated even in very deep networks. Second, ResNets are robust to layer removal at training time \cite{HuangSLSW16} and at inference time \cite{VeitWB16} implying that the operations applied by a single layer are only a small modification to identity operation. Third, ResNets are robust to layers permutation \cite{VeitWB16}, suggesting that neighboring layers perform similar operations. These characteristics are not shared with traditional CNNs and researchers have attempted to propose possible explanations on the internal behaviors and mechanisms with ResNet-like models. Recently, two possible explanations of ResNets-like models have emerged. The first, explains the behavior of ResNets in terms of an \textit{embedding of relatively shallow networks} \cite{VeitWB16}. The second, suggests that ResNets perform \textit{iterative estimation}, where the input to the model is iteratively modified by small transformations \cite{Greff16,LiaoP16}.

In recent years, state of the art segmentation methods for medical images have been based on Fully Convolutional Networks (FCNs)~\cite{long_shelhamer_fcn, RonnebergerFB15}. While CNNs typically consist of a contracting path composed of convolutional, pooling and fully connected layers, FCNs add an expanding path composed of transposed convolutions or unpooling layers. The expanding path recovers spatial information by merging features skipped from the various resolution levels on the contracting path. Variants of these skip connections are proposed in the literature. In \cite{long_shelhamer_fcn}, upsampled feature maps are summed with feature maps skipped from the contractive path, while \cite{RonnebergerFB15} concatenate them and add convolutions and non-linearities between each upsampling step. These skip connections have been shown to help recover the full spatial resolution at the network's output, making fully convolutional methods suitable for semantic segmentation. Since traditional FCNs are an extension of CNNs, they can be explained by the representation learning perspective on deep learning.

\begin{figure}[t!]
\centering
\subfigure[Traditional pipeline]{\includegraphics[width=0.5\textwidth]{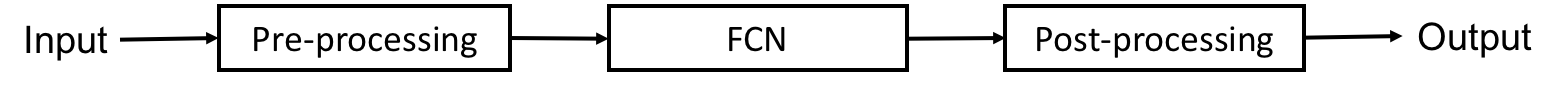}\label{fig:01}}\hfill
\subfigure[Our pipeline]{\includegraphics[width=0.35\textwidth]{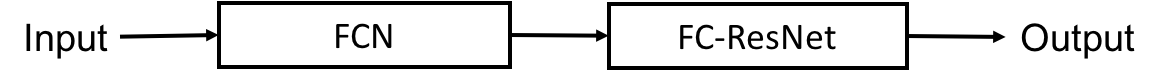}\label{fig:02}}\hfill
\caption{Traditional pipeline for medical image segmentation (a) and proposed pipeline for medical image segmentation (b).}
\label{fig:pipeline}
\vspace{-.5cm}
\end{figure}

Although deep learning methods have proved their potential in medical image segmentation, their performance strongly depends on the quality of pre-processing and post-processing steps \cite{HavaeiGLJ16}. Thus, traditional image segmentation pipelines based on FCNs are often complemented by pre-processing and post-processing blocks (see Figure\ref{fig:01}). Pre-processing methods vary among different imaging modalities and can include operations like standardization, histogram equalization, value clipping or range normalization (e.g. dividing by maximum intensity value). The tools of choice for post-processing are either based on Conditional Random Fields \cite{Krahenbuhl11} to account for spatial consistency of the output prediction or on morphological operations to clean the output prediction.

In \cite{DrozdzalVCKP16}, the Fully Convolutional Residual Networks (FC-ResNets) are introduced. FC-ResNets incorporate additional shortcut paths and, thus, increase the number of connections within a segmentation network. These additional shortcut paths have been shown not only to improve the segmentation accuracy but also help the network optimization process, resulting in faster convergence of the training. Since FC-ResNets are an extension of ResNets, their behavior should be interpreted in terms of iterative estimation or embedding of relatively shallow networks. Moreover, not surprisingly, these FC-ResNets are more susceptible to image pre-processing than FCNs, their performance is highly dependent on proper data preparation (e.g. data standardization or range normalization). 

In this paper, we take advantage of recent advances in the understanding of both CNNs as well as ResNets and propose a new medical image segmentation pipeline. We use a FCN to obtain pre-normalized images, which are then iteratively refined by means of a FC-ResNet to generate a segmentation prediction (see Figure \ref{fig:02}). 

Thus, in our pipeline, the FCN can be thought of as a pre-processor that is learnt by means of back-propagation, and FC-ResNet can be thought as a powerful classifier that is an ensemble of exponential number of shallow models. This small modification to current segmentation pipelines allows to remove hand-crafted data pre-processing and to build end-to-end systems trained with back-propagation that achieve surprisingly good performance in segmentation tasks for bio-medical images. Our pipeline reaches state-of-the art for 2D methods on electron microscopy (EM) ISBI benchmark dataset \cite{EM_data} and outperforms both standard FCN \cite{long_shelhamer_fcn,RonnebergerFB15} and FC-ResNet \cite{DrozdzalVCKP16} on in-house CT liver lesion segmentation dataset. Moreover, while applying our 2D pipeline off-the-shelf on a challenging 3D MRI prostate segmentation challenge, we reach results that are competitive even when compared to 3D methods.

Thus, the contributions in this paper can be summarized as follows:
\begin{itemize}
\item We combine Fully Convolutional Residual Networks with Fully Convolutional Networks (Section \ref{sec:Method}).
\item We show that a very deep network without any hand-designed pre- or post-processing achieves state-of-the-art performance on the challenging EM ISBI benchmark \cite{EM_data} (Section \ref{sec:EM_data}).
\item We show that our pipeline outperforms other common FCN-based segmentation pipelines on our in-house CT liver lesion dataset (Section \ref{sec:liver_data}).
\item We show that our 2D pipeline can be applied to untreated MR images reaching competitive results on challenging 3D MRI prostate segmentation task, outperforming many 3D based approaches (Section \ref{sec:prostate_data}).
\item We show that a FCN based pre-processor normalizes the data to the values adequate for FC-ResNet (Section \ref{sec:normalizatoin}).
\end{itemize}

\section{Background}
\label{sec:Background}

Recent advances in medical image segmentation often involve convolutional networks. Most of these state-of-the-art approaches are based on either variants of FCNs (FCN8 or UNet) or CNN architectures. FCN architectures process the input image end-to-end and provide a full resolution segmentation map, whereas CNN variants are applied to input patches and aim to solely classify the central pixel of each patch. In many cases, the application of an FCN/CNN is preceded by a pre-processing step and followed by a post-processing step. The former aims to account for the variability in the input images, whereas the latter helps refine the predictions made by the FCN/CNN.

In the remainder of this section, we review segmentation pipelines in medical imaging based on deep neural networks, for different imaging modalities and organs, with a particular focus on pre-processing and post-processing steps proposed to normalize and regularize data, respectively.

\textbf{Electron Microscopy (EM)}. EM is widely used to study synapses and other sub-cellular structures in the mammalian nervous system. EM data is the basis of two medical imaging segmentation challenges: 2D \cite{EM_data} and 3D \cite{EM3D}. The core of the best performing methods is based on a patch-based CNN \cite{Ciresan}, FCN8 \cite{0011QCH16}, UNet \cite{RonnebergerFB15} and FC-ResNets \cite{DrozdzalVCKP16, Quan2016, Fakhry2016}. The methods account for gray scale value variability by employing data augmentation (e.g. intensity shifts \cite{RonnebergerFB15,Quan2016}) or data pre-processing (e.g standardization \cite{DrozdzalVCKP16} or rescaling \cite{Quan2016}). As for post-processing, a variety of FCN prediction refinement techniques have been proposed. In \cite{Beier2016}, a minimum cost multi-cut approach is introduced, in \cite{Fakhry2016, 0011QCH16}, watershed algorithm is used, while in \cite{Quan2016}, median filtering is employed to improve EM segmentation results.

\begin{figure*}[t!]
\centering
\hspace*{\fill}%
\subfigure[FC-ResNet]{\includegraphics[width=0.4\textwidth]{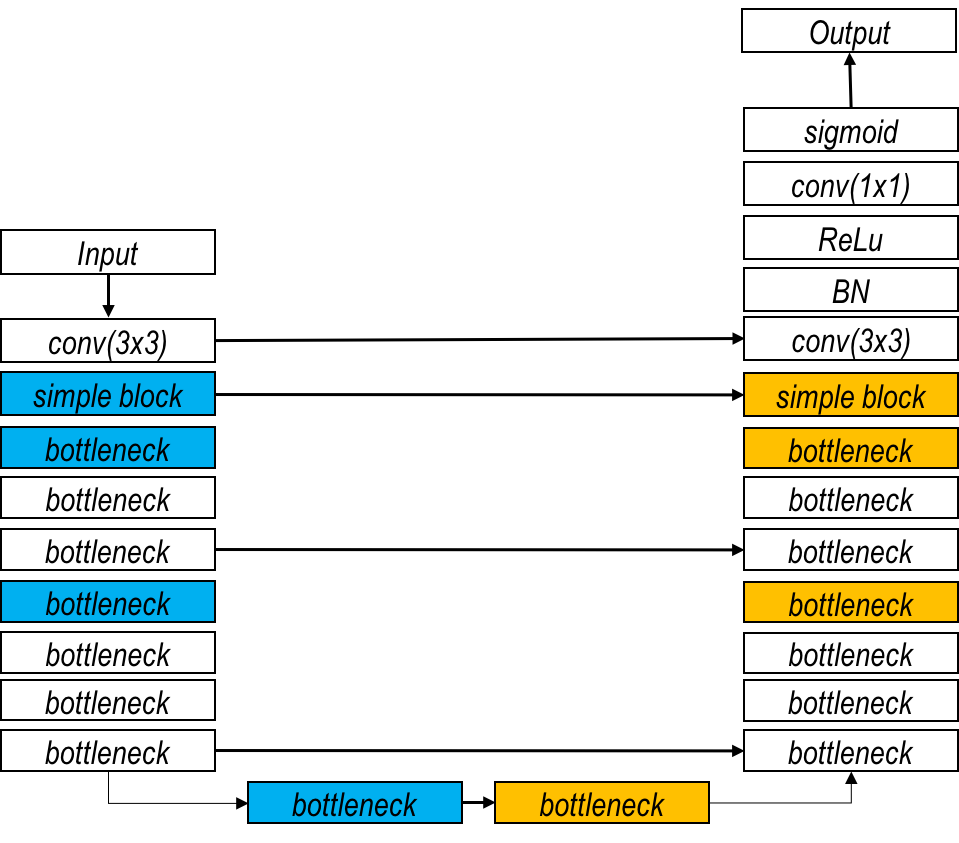}\label{fig:1}}\hfill
\subfigure[Bottleneck block]{\includegraphics[width=0.135\textwidth]{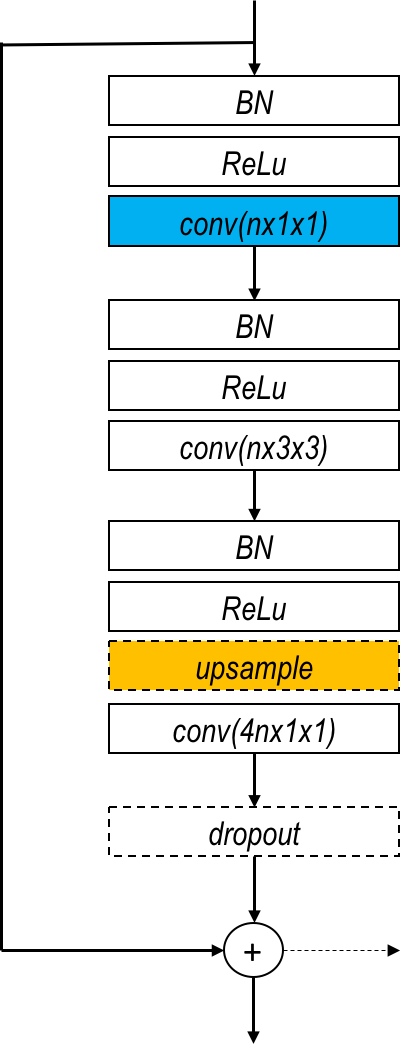}\label{fig:2}}\hfill
\subfigure[Simple block]{\includegraphics[width=0.135\textwidth]{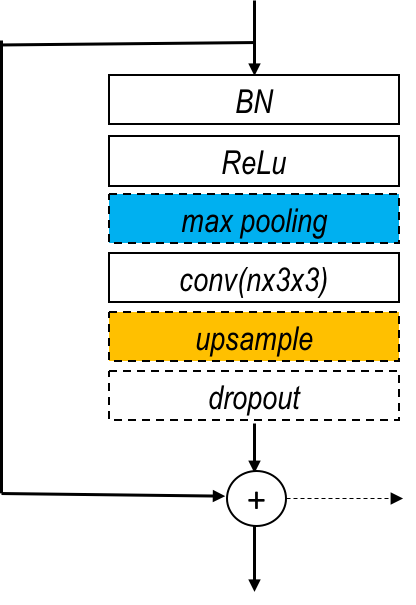}\label{fig:4}}
\hspace*{\fill}%
\caption{An example of residual network for image segmentation. (a) Residual Network with long skip connections built from bottleneck blocks, (b) bottleneck block and (c) simple block. Blue color indicates the blocks where a downsampling is optionally performed, yellow color depicts the (optional) upsampling blocks, dashed arrow in figures (b) and (c) indicates possible long skip connections. Note that blocks (b) and (c) can have a dropout \cite{Srivastava:2014} layer (depicted with dashed line rectangle). The downsampling operation is performed with strided convolution in the bottleneck block and with maxpooling in the simple block.}
\label{fig:ResUNet}
\end{figure*}


\textbf{Magnetic Resonance Imaging (MRI).} MRI is used for detection, classification of lesions and tumor staging. As in many medical imaging applications, deep learning methods are nowadays playing an important role in the segmentation of lesions and organs from MRI scans, consistently improving state-of-the-art performance. Patch-based CNN architectures have been used to segment brain tumors \cite{KamnitsasLNSKMR16,Pandian116} and MS lesions \cite{BirenbaumG16}. Variants of FCN8 and UNets have also been applied to segment brain tumors \cite{Chang16,Lun16,McKinley16,Casamitjana16,Zhao16}, whereas FC-ResNets have been successfully employed to segment white and gray matter in the brain \cite{Chen16_vox} and prostate \cite{Yu17,MilletariNA16}. To account for 3D information, some of the proposed architectures apply convolutions in a 3D fashion \cite{Chen16_vox,Casamitjana16,Pandian116,KamnitsasLNSKMR16,MilletariNA16} or make use of recurrent neural networks (LSTMs or GRUs) \cite{Andermatt2016,StollengaBLS15}. As in the previous modalities, pre-processing techniques are applied to the input of the network. MRI pre-processing techniques include intensity normalization \cite{BirenbaumG16,Chen16_vox}, rescaling \cite{Chang16}, standardization \cite{KamnitsasLNSKMR16,Pandian116,Casamitjana16,Zhao16,Chen16_vox,Andermatt2016,StollengaBLS15,Yu17}, N4 bias correction \cite{Pandian116,Zhao16,BirenbaumG16,StollengaBLS15} and histogram equalization \cite{Chang16}. Again, post-processing involving morphological operations \cite{Pandian116,Zhao16}, conditional random fiels (CRFs) \cite{KamnitsasLNSKMR16,Zhao16} and interpolation \cite{Andermatt2016} are often used to refine segmentation maps.

\textbf{Computed Tomography (CT)}. Finally, CT scans are widely used by clinicians to detect a large variety of lesions in different organs. Patch-based CNNs have been used in the literature to segment pathological kidneys \cite{Zheng16,Thong16}, liver tumors \cite{VivantiEJLKS15}, pancreas \cite{RothLFSLTS15} and urinary bladder \cite{Cha16}. FCN8 and UNet variants have also been tried on CT scans, e.g. to segment liver tumors \cite{Ben-CohenDKAG16,ChristEETBBRAHD16}. In many cases, CT scans are pre-processed by standardizing \cite{Thong16,Li15,Cha16}, clipping \cite{ChristEETBBRAHD16}, applying Gaussian smoothing \cite{Li15,Cha16} or histogram equalization \cite{ChristEETBBRAHD16} to the input. Gaussian noise injection has also been explored as part of data augmentation to account for noise level variability in the CT scans \cite{ChristEETBBRAHD16}. As in the other modalities, the most frequently used post-processing methods include morphological operations \cite{Thong16,VivantiEJLKS15,RothLFSLTS15}, CRFs \cite{ChristEETBBRAHD16} and level sets \cite{Cha16} to refine the segmentation proposals.

\section{Method}
\label{sec:Method}

In this section we explain our segmentation pipeline. As explained in Section \ref{sec:Intro}, our approach combines a FCN model with a FC-ResNet model (see Figure \ref{fig:02}). The goal of FCN in our pipeline is to pre-process the image to a format that can be iteratively refined by FC-ResNet. In the remainder of this section, we describe the architecture of our FCN-based pre-processor (see Subsection \ref{sec:FCN_pre_processor}), our FC-ResNet (see Subsection \ref{sec:FC-ResNet}) as well as the loss function used to train our pipeline (see Subsection \ref{sec:loss}).

\subsection{Fully Convolutional pre-processor}
\label{sec:FCN_pre_processor}
Our FCN takes as input a raw image of size $N \times N \times 1$ (e.g. CT scan slice or EM image) without applying any pre-processing and outputs a processed $N \times N \times 1$ feature map. The FCN pre-processor architecture is described as a variation of the UNet model from \cite{RonnebergerFB15} (see Table \ref{tab:FCN} for details). The contracting path is built by alternating convolutions and max pooling operations, whereas the expanding path is built by alternating convolutions and repeat operations. The expanding path recovers spatial information, lost in pooling operations, by concatenating the corresponding feature maps from the contracting path. In total, the model has 4 pooling operations and 4 repeat operations. All $3 \times 3$ convolutions are followed by ReLU non-linearity, while no non-linearity follows the $1 \times 1$ and $2 \times 2$ convolutions. In our experiments, we reduce the number of feature maps by a factor of 4 when compared to the original UNet (e. g. our first layer has 16 feature maps instead of 64 as in the original model). This step significantly reduces the memory foot-print of the FCN-based pre-processor. Thus, our UNet-like pre-processor has $1.8$ million trainable parameters (vs 33 million in the original implementation of \cite{RonnebergerFB15}). Note that the UNet could potentially be replaced by other FCN models (e.g. FCN8 \cite{long_shelhamer_fcn}).

\subsection{Iterative Estimation with FC-ResNets}
\label{sec:FC-ResNet}

ResNets \cite{HeZRS15} introduce a residual block that sums the identity mapping of the input to the output of a layer allowing for the reuse of features and permitting the gradient to flow directly to earlier layers. The resulting output $x_{\ell}$ of the $\ell^{th}$ block becomes
\begin{equation}
x_\ell = H_\ell(x_{\ell-1}) + x_{\ell-1}, 
\end{equation}
where $H$ is defined as the repetition (2 or 3 times) of a block composed of Batch Normalization (BN) \cite{IoffeS15} followed by ReLU and a convolution.

The FC-ResNet model extends ResNets to be fully convolutional by adding an expanding (upsampling) path (Figure \ref{fig:1}). Spatial reduction is performed along the contracting path (left) and expansion is performed along the expanding path (right). As in \cite{long_shelhamer_fcn} and \cite{RonnebergerFB15}, spatial information lost along the contracting path is recovered in the expanding path by skipping equal resolution features from the former to the latter. Similarly to the identity connections in ResNets, the skipped features coming from the contracting path are summed with the ones in the expanding path. 

Following the spirit of ResNets, FC-ResNets are composed of two different types of blocks: simple blocks and bottleneck blocks, each composed of at least one batch normalization followed by a non-linearity and one convolution (see Figure \ref{fig:2}-\ref{fig:4}). These blocks can maintain the spatial resolution of their input (marked white in Figure \ref{fig:2}-\ref{fig:4}), perform spatial downsampling (marked blue in Figure \ref{fig:2}-\ref{fig:4}) or spatial upsampling (marked yellow in Figure \ref{fig:2}-\ref{fig:4}). As in ResNets, bottleneck blocks are characterized by their $1 \times 1$ convolutions, which are responsible for reducing and restoring the number of feature maps and thus, aim to mitigate the number of parameters of the model.

The detailed description of FC-ResNet architecture used in our pipeline is shown in Table \ref{tab:FC-ResNet}. The contracting path contains 5 downsampling operations, one $3\times3$ convolution, one simple block and 21 bottleneck blocks. The contracting path is followed by 3 bottleneck blocks, which precede the expanding path. The expanding path contains 5 upsampling operations, 21 bottleneck blocks, one simple block and one last $3\times3$ convolution. This FC-ResNet has a total of $11$ millions trainable parameters.

To sum up, our model is composed of a UNet-like model followed by a FC-ResNet. The UNet-like model is composed of 23 convolutional layers and $1.8$ million trainable parameters. Our FC-ResNet has 140 convolutional layers and $11$ millions of trainable parameters. Thus, our full segmentation pipeline has $12.8$ millions of trainable parameters.

\begin{table}[t!]
\begin{center}
\begin{tabular}{||c |c |c |c |c||} 
\hline
\rotatebox{0}{\thead{Layer \\ Name}} & \rotatebox{0}{\thead{Block \\ Type}} & \rotatebox{0}{\thead{Output \\ Resolution}} & \rotatebox{0}{\thead{Output \\ Width}} & \rotatebox{0}{\thead{Repetition \\ Number}} \\ [0.5ex] 
\hline\hline
Input & - & $512 \times 512$ & 1 & -\\
\hline
Down 1 & conv $3 \times 3$ & $512 \times 512$ & 16 & 2\\
\hline
Pooling 1 & maxpooling & $256 \times 256$ & 16 & 1 \\ 
\hline
Down 2 & conv $3 \times 3$ & $256 \times 256$ & 32 & 2\\
\hline
Pooling 2 & maxpooling & $128 \times 128$ & 32 & 1 \\
\hline
Down 3 & conv $3 \times 3$ & $128 \times 128$ & 64 & 2 \\
\hline
Pooling 3 & maxpooling & $64\times 64$ & 64 & 1 \\
\hline
Down 4 & conv $3 \times 3$ & $64\times 64$ & 128 & 2 \\
\hline
Pooling 4 & maxpooling & $32\times 32$ & 128 & 1 \\
\hline
Across & conv $3 \times 3$ & $32\times 32$ & 256 & 2 \\
\hline
Up 1 & upsampling & $64 \times 64$ & 256 & 1 \\
\hline
Merge 1 & concatenate & $64 \times 64$ & 384 & 1 \\
\hline
Up 2 & conv $2 \times 2$ & $64 \times 64$ & 128 & 1\\
\hline
Up 3 & conv $3 \times 3$ & $64 \times 64$ & 128 & 2 \\
\hline
Up 4 & upsampling & $128 \times 128$ & 128 & 1 \\
\hline
Merge 2 & concatenate & $128 \times 128$ & 192 & 1 \\
\hline
Up 5 & conv $2 \times 2$ & $128 \times 128$ & 64 & 1\\
\hline
Up 6 & conv $3 \times 3$ & $128 \times 128$ & 64 & 2 \\
\hline
Up 7 & upsampling & $256 \times 256$ & 64 & 1 \\
\hline
Merge 3 & concatenate & $256 \times 256$ & 96 & 1 \\
\hline
Up 8 & conv $2 \times 2$ & $256 \times 256$ & 32 & 1\\
\hline
Up 9 & conv $3 \times 3$ & $256 \times 256$ & 32 & 2 \\
\hline
Up 10 & upsampling & $512 \times 512$ & 32 & 1 \\
\hline
Merge 4 & concatenate & $512 \times 512$ & 48 & 1 \\
\hline
Up 11 & conv $2 \times 2$ & $512 \times 512$ & 16 & 1\\
\hline
Up 12 & conv $3 \times 3$ & $512 \times 512$ & 16 & 2 \\
\hline
Output & conv $3 \times 3$ & $512 \times 512$ & 1 & 1\\
\hline
\end{tabular}
\end{center}
\caption{Detailed FCN architecture used as a pre-processor in the experiments. Output resolution indicates the spatial resolution of feature maps for an input of size $512\times512$ and output width represents the feature map dimensionality. Repetition number indicates the number of times the layer is repeated.}
\label{tab:FCN}
\vspace{-.5cm}
\end{table}

\begin{table}[t!]
\begin{center}
\begin{tabular}{||c |c |c |c |c||} 
\hline
\rotatebox{0}{\thead{Layer \\ Name}} & \rotatebox{0}{\thead{Block \\ Type}} & \rotatebox{0}{\thead{Output \\ Resolution}} & \rotatebox{0}{\thead{Output \\ Width}} & \rotatebox{0}{\thead{Repetition \\ Number}} \\ [0.5ex] 
\hline\hline
Down 1 & conv $3 \times 3$ & $512 \times 512$ & 32 & 1\\
\hline
Down 2 & simple block & $256 \times 256$ & 32 & 1 \\ 
\hline
Down 3 & bottleneck & $128 \times 128$ & 128 & 3\\
\hline
Down 4 & bottleneck & $64 \times 64$ & 256 & 8 \\
\hline
Down 5 & bottleneck & $32 \times 32$ & 512 & 10 \\
\hline
Across & bottleneck & $32\times 32$ & 1024 & 3 \\
\hline
Up 1 & bottleneck & $64 \times 64$ & 512 & 10 \\
\hline
Up 2 & bottleneck & $128 \times 128$ & 256 & 8\\
\hline
Up 3 & bottleneck & $256 \times 256$ & 128 & 3 \\
\hline
Up 4 & simple block & $512 \times 512$ & 32 & 1\\
\hline
Up 5 & conv 3x3 & $512 \times 512$ & 32 & 1 \\
\hline
Classifier & conv $1 \times 1$ & $512 \times 512$ & 1 & 1\\ 
\hline
\end{tabular}
\end{center}
\caption{Detailed FC-ResNet architecture used in the experiments. Output resolution indicates the spatial resolution of feature maps for an input of size $512\times512$ and output width represents the feature map dimensionality. Repetition number indicates the number of times the block is repeated. For the definition of block types, please refer to Figures \ref{fig:2} and \ref{fig:4}.}
\label{tab:FC-ResNet}
\vspace{-.5cm}
\end{table}

\subsection{Dice Loss}
\label{sec:loss}
We train our model using the Dice loss ($L_{Dice}$) computed per batch:
\begin{equation}
L_{Dice} = -\frac{2 \sum_i o_i y_i}{\sum_i o_i + \sum_i y_i},
\end{equation}
where $o_i \in [0, 1]$ represents the $i^{th}$ output of the last network layer (sigmoid output) and $y_i \in \{0, 1\}$ represents the corresponding ground truth label. Note that the minimum value of  the Dice loss is $-1$.

The reason for using the Dice loss over traditional binary crossentropy is two-fold. First, the Dice coefficient is of the common metrics to assess medical image segmentation accuracy, thus, it is natural to optimize it during training. Second, as pointed out in \cite{MilletariNA16}, Dice loss is well adapted to the problems with high imbalance between foreground and background classes as it does not require any class frequency balancing.

\section{Experiments}
\label{sec:Results}
In this section, we present experimental results of the proposed pipeline for image segmentation. First, we show that our method achieves state-of-the-art results among all published 2D methods on challenging EM benchmark \cite{EM_data}. Second, we compare our pipeline with standard FCNs \cite{long_shelhamer_fcn,RonnebergerFB15} and with FC-ResNets \cite{DrozdzalVCKP16} on a dataset of CT scans of liver lesions, with 135 manually annotated scans. Finally, we show the normalization effect of the FCN-based pre-processing module on both datasets.

In our experiments, we train with early stopping at the highest Dice value on the validation set with patience of $50$ epochs. We implemented our model in the Keras framework \cite{chollet2015keras} using the Theano backend \cite{theano}.

\subsection{Electron Microscopy dataset}
\label{sec:EM_data}

The EM training dataset consists of $30$ images ($512 \times 512$ pixels) assembled from serial section transmission electron microscopy of the Drosophila first instar larva ventral nerve cord. The test set is a separate set of $30$ images, for which labels are not provided. 

The official metrics used in this dataset are: Maximal foreground-restricted Rand score after thinning ($V_{rand}$) and maximal foreground-restricted information theoretic score after thinning ($V_{info}$) with ($V_{rand}$), being used to order the entries in the leader board. For a detailed description of the metrics, please refer to \cite{EM_data}.

During training, we augmented the dataset using random flipping (horizontal and vertical), sheering (with maximal range of $0.41$), rotations (with maximal range of $25$), random cropping ($256\times256$) and spline warping. We used the same spline warping strategy as \cite{RonnebergerFB15}. Thus, our data augmentation is similar to the one published in \cite{DrozdzalVCKP16}. We trained the model with RMSprop \cite{Tieleman2012} with an initial learning rate of $0.001$, a learning rate decay of $0.001$ and a batch size of $8$. We used weight decay of $0.0001$. For each training, the model with the best validation Dice was stored. In total, we trained 10 models and averaged their outputs at the test time. Each time the model was trained, we randomly split $30$ images into $24$ training images and $6$ validation images. 

\begin{figure*}[t!]
\centering
\hspace*{\fill}%
\subfigure[Input image]{\includegraphics[width=0.24\textwidth]{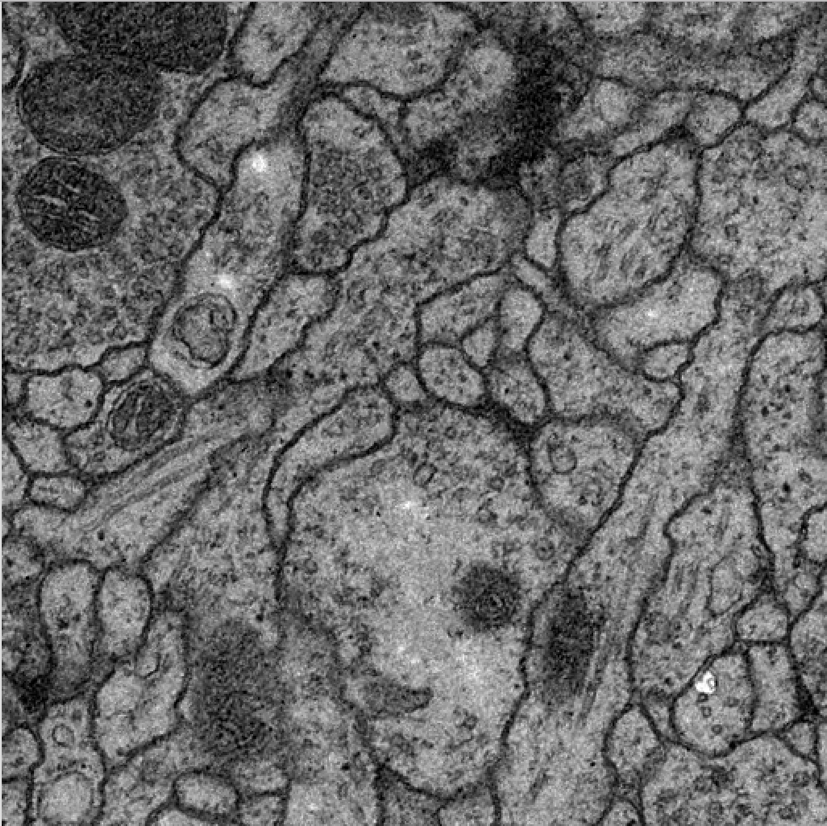}\label{fig:EM1}}\hfill
\subfigure[FC-ResNet with dropout at test time \cite{DrozdzalVCKP16}]{\includegraphics[width=0.24\textwidth]{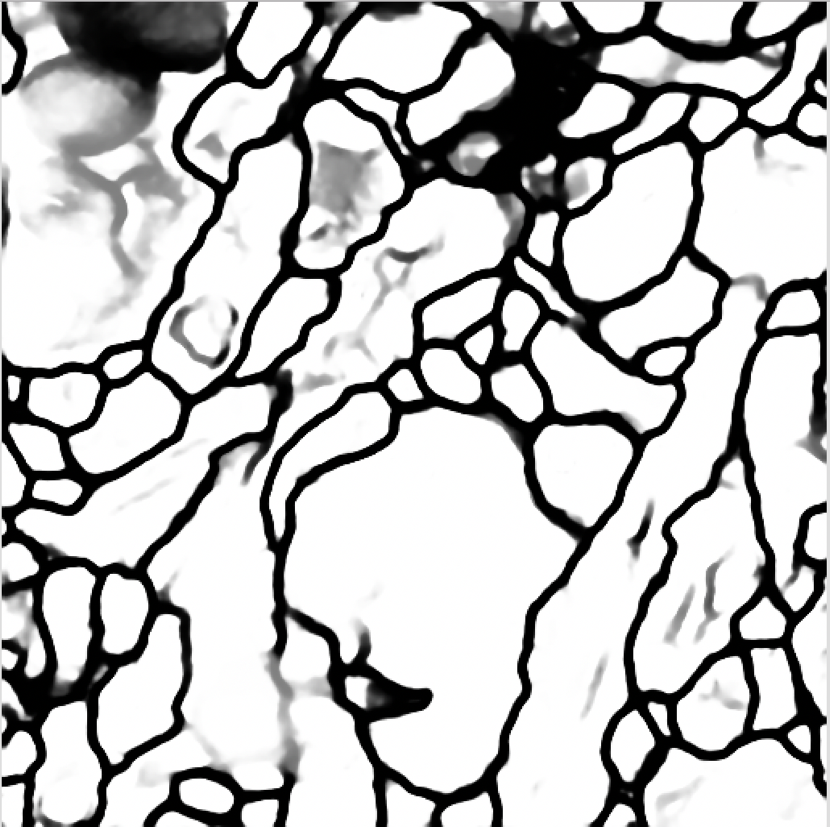}\label{fig:EM2}}\hfill
\subfigure[Segmentation result of our pipeline]{\includegraphics[width=0.24\textwidth]{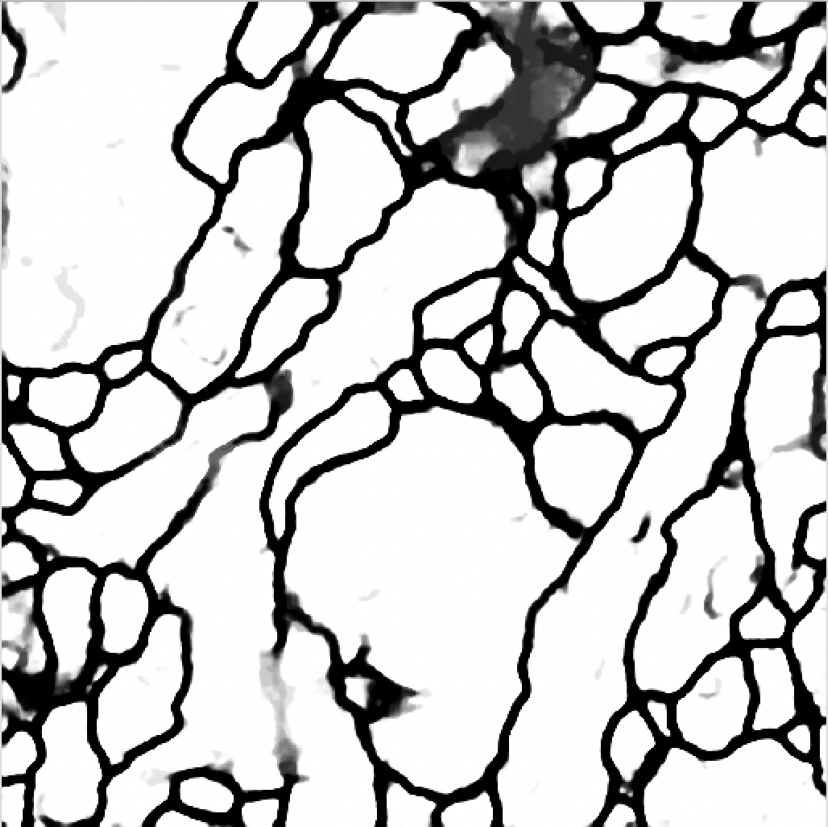}\label{fig:EM3}}
\hspace*{\fill}%
\caption{Qualitative segmentation results for EM data. (a) An image from the test set, (b) prediction of FC-ResNet with standarization as a pre-porcessor and dropout at test time and (c) prediction obtained by our pipeline with FCN-based processor followed by FC-ResNet.}
\label{fig:EM}
\end{figure*}

\begin{table*}[t!]
\begin{center}
\begin{tabular}{||c |c |c ||c |c |c |c ||} 
\hline
\thead{Method} & \thead{\boldmath$V_{rand}$} & \thead{\boldmath$V_{info}$}  & \thead{post-processing} & \thead{pre-processing} & \thead{average over} & \thead{parameters [M]}\\ [0.5ex] 
\hline\hline
FusionNet \cite{Quan2016} & 0.978 & \textbf{0.990} & YES & YES & 8 &  31 \\
\hline
CUMedVision \cite{0011QCH16} & 0.977 & 0.989 & YES & NO & 6  &  8 \\
\hline
Unet \cite{RonnebergerFB15} & 0.973 & 0.987 & NO &  YES & 7 & 33  \\ 
\hline
FC-ResNet \cite{DrozdzalVCKP16} & 0.969 & 0.986 & NO & YES & Dropout & 11 \\
\hline
\hline
Ours & \textbf{0.981} & 0.988 & NO & NO & 10 & 13 \\%
\hline
\end{tabular}
\end{center}
\caption{Comparison to published FCN results for EM dataset. $V_{rand}$ is the measure used to rank the submissions. For full ranking of all submitted methods please refer to the challenge website: \protect\url{http://brainiac2.mit.edu/isbi_challenge/leaders-board-new}.}
\label{tab:score_FCN}
\vspace{-.5cm}
\end{table*}

\begin{table}[t!]
\begin{center}
\begin{tabular}{||c |c |c |c ||} 
\hline
\thead{Method} & \thead{\boldmath$V_{rand}$} & \thead{\boldmath$V_{info}$} & \thead{2D/3D}\\ [0.5ex] 
\hline\hline
IAL \cite{Beier2016} & 0.980 & 0.988 & 2D\\
\hline
FusionNet \cite{Quan2016} & 0.978 & \textbf{0.990} & 2D\\
\hline
CUMedVision \cite{0011QCH16} & 0.977 & 0.989 & 2D\\
\hline
Unet \cite{RonnebergerFB15} & 0.973 & 0.987 & 2D\\ 
\hline
IDSIA \cite{Ciresan}  & 0.970 & 0.985 & 2D\\
\hline
motif \cite{Wu15f}  & 0.972 & 0.985 & 2D\\
\hline
SCI \cite{Liu201488}  & 0.971 & 0.982 & 3D\\
\hline
optree-idsia\cite{Uzunbaş2014}  & 0.970 & 0.985 & 2D \\
\hline
FC-ResNet \cite{DrozdzalVCKP16} & 0.969 & 0.986 & 2D\\
\hline
PyraMiD-LSTM\cite{StollengaBLS15}  & 0.968 & 0.983 & 3D\\
\hline
\hline	
Ours & \textbf{0.981} & 0.988 & 2D\\%
\hline
\end{tabular}
\end{center}
\caption{Comparison to published entries for EM dataset. $V_{rand}$ is the measure used to order the submissions. For full ranking of all submitted methods please refer to the challenge website: \protect\url{http://brainiac2.mit.edu/isbi_challenge/leaders-board-new}.}
\label{tab:score}
\vspace{-.5cm}
\end{table}

The comparison of our method to other FCNs is shown in Table \ref{tab:score_FCN}. Our pipeline outperforms all other fully convolutional approaches when looking at the primary metric ($V_{rand}$ score), improving it by 0.003 over the second best fully convolutional approach (FusionNet \cite{Quan2016}). When comparing our pipeline (FCN pre-processing followed by FC-ResNet) to the pipeline of \cite{DrozdzalVCKP16} (standardization pre-processing followed by FC-ResNet), we observe a significant increase in performance. The same happens when comparing our pipeline to FusionNet \cite{Quan2016} (rescaling pre-processing followed by FC-ResNet). It is worth noting that FusionNet applies intensity shifts with Gaussian noise as data augmentation to account for input variability. All these results suggest that FC-ResNet is better complemented by an FCN pre-processor than by traditional pre-processors such as standardization or rescaling. It is worth noting that our pipeline has only slightly more parameters than \cite{DrozdzalVCKP16} and significantly fewer parameters than \cite{Quan2016}. There are two additional fully convolutional approaches submitted for this dataset: UNet \cite{RonnebergerFB15} and CUMedVision \cite{0011QCH16}, the latter being based on FCN8 \cite{long_shelhamer_fcn}.  The pre-processing used on this dataset varies from range normalization to values from 0 to 1 \cite{Quan2016} to data standardization \cite{DrozdzalVCKP16}. For post-processing, median filter \cite{Quan2016} and watershed algorithm \cite{0011QCH16} are used. All methods use either prediction or model averaging at test time.

Table \ref{tab:score} compares our method to other published entries for EM dataset. As shown in the table, we report the highest $V_{rand}$ score (at the moment of writing the paper), when compared to all published 2D methods. The second best entry is IAL \cite{Beier2016}, which introduces a graph-cut post-processing method to refine FCN predictions. In \cite{Beier2016}, a 2D approach was tested resulting in $0.980$ $V_{rand}$ score and $0.988$ $V_{info}$ score. However, IAL leaderboard entrance incorporates 3D context information and achieves an improvement of $0.003$ in $V_{rand}$ score and $0.001$ in $V_{info}$ score\footnote{Personal communication with the authors.} w.r.t. IAL 2D entry. 

Finally, we show some qualitative analysis of our method in Figure \ref{fig:EM}, where we display a prediction of a single test frame for two different pipelines: FC-ResNet from \cite{DrozdzalVCKP16} in Figure \ref{fig:EM2} and of our pipeline in Figure \ref{fig:EM3}. The white color in the prediction images correspond to the cell class and the black color correspond to the cell membrane class. The different degrees of gray correspond to regions in which the model is (more or less) uncertain about the class assignment. The difference is especially visible when comparing our model to FC-ResNet, we can see that the predictions are sharper and clearer, suggesting that the FCN-pre-processor has properly prepared the image for FC-ResNet.

\subsection{Liver lesion dataset}
\label{sec:liver_data}

Our in-house dataset consists of abdominal contrast-enhanced CT-scans from patients diagnosed with colorectal metastases (CRM). All images are $512 \times 512$ with a pixel size varying from $0.53$ to $1.25$ mm and a slice thickness varying from $0.5$ to $5.01$ mm. The pixel intensities vary between $-3000$ and $1500$. For each volume, CRM were segmented manually using MITK Workbench \cite{MITK} by medical students and reviewed by professional image analysts, resulting in 135 CT scans with manually segmented CRM. In addition to that, manual liver segmentation was provided for 58 of the 135 CT scans. This data was collected with the specific goal of segmenting lesions \emph{only} within the liver.

We split the dataset to have $77$ training images with CRM segmented, $28$ validation images with liver and CRM segmented and 30 test images with liver and CRM segmented. Note that for the training set we do not have liver segmentations available. The liver segmentations of both the validation and the test sets are used to limit the lesion segmentation evaluation only to the liver, treating the rest of the image as a void class.

During the training, we randomly crop a 2D $128 \times 128$ pixel patch containing CRM from a CT scan. We train the model with RMSprop \cite{Tieleman2012} with an initial learning rate of $0.001$ and a learning rate decay of $0.001$. We use weight decay of $0.0001$ in the pre-processor and $0.0005$ in FC-ResNet. 

\begin{table*}[t!]
\begin{center}
\begin{tabular}{||c |c ||c |c |c ||c |c |c ||} 
\hline
\multirow{2}{*}{\thead{Method}} &
\multirow{2}{*}{\thead{parameters [M]}} &
      \multicolumn{3}{c||}{\thead{Validation}} &
      \multicolumn{3}{c||}{\thead{Test}} \\
    & & loss & $Dice_{lesion}$ & $Dice_{liver}$ & loss & $Dice_{lesion}$ & $Dice_{liver}$ \\
\hline\hline
FCN8 \cite{long_shelhamer_fcn} & 128 & -0.419 & 0.589 & 0.994 & -0.437 & 0.535 & 0.989 \\
\hline
Unet \cite{RonnebergerFB15} & 33 & -0.451 & 0.553 & 0.994 & -0.396 & 0.570 & 0.990 \\ 
\hline
FC-ResNet \cite{DrozdzalVCKP16} & 11 & -0.223 & 0.551 & 0.993 & -0.224 & 0.617 & 0.990 \\
\hline
\hline
Ours & 13 & \textbf{-0.795} & \textbf{0.771} & \textbf{0.997} & \textbf{-0.796} & \textbf{0.711} & \textbf{0.993} \\%
\hline
\end{tabular}
\end{center}
\caption{Results on the liver lesion dataset for both validation and test sets.}
\label{tab:score_liver}
\vspace{-.5cm}
\end{table*}

We follow the same training procedure for all FCN methods: FCN8 \cite{long_shelhamer_fcn}, UNet \cite{RonnebergerFB15} and FC-ResNets \cite{DrozdzalVCKP16}. Results are reported in Table \ref{tab:score_liver}. All models were trained with the Dice loss with batch size of $20$ at training time and $1$ at validation and test time. Our approach outperforms other methods on this challenging dataset, achieving the best validation loss of $-0.795$ and lesion validation Dice of $0.771$. The second best validation loss was obtained by the UNet model ($-0.451$) and the second best lesion validation Dice was obtained by the FCN8 model ($0.589$). Moreover, our model generalizes well on test set reaching a loss of $-0.796$ and a Dice of $0.711$.

\begin{figure*}[t!]
\centering
\subfigure{\includegraphics[width=0.2\textwidth]{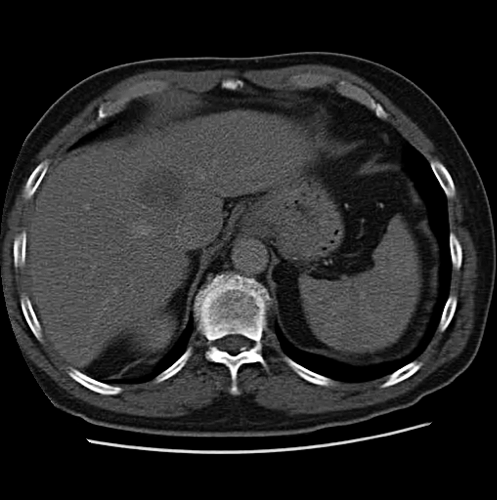}\label{fig:liver1}}\hfill
\subfigure{\includegraphics[width=0.2\textwidth]{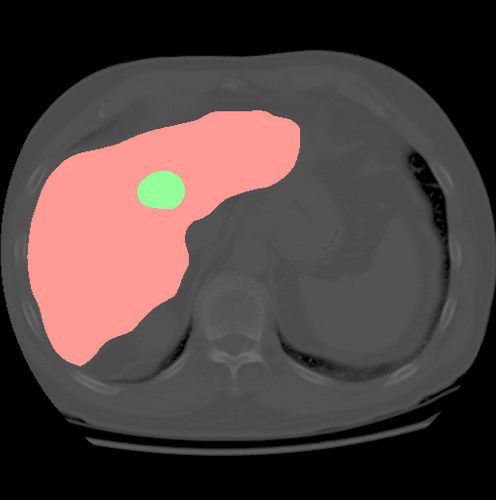}\label{fig:liver2}}\hfill
\subfigure{\includegraphics[width=0.132\textwidth]{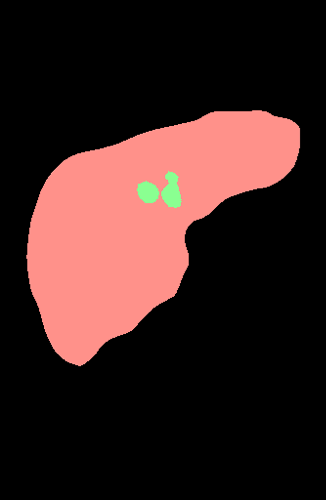}\label{fig:liver3}}\hfill
\subfigure{\includegraphics[width=0.132\textwidth]{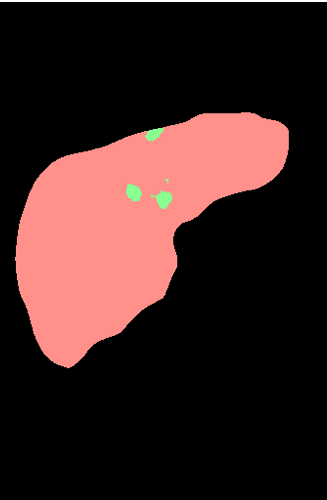}\label{fig:liver4}}\hfill
\subfigure{\includegraphics[width=0.132\textwidth]{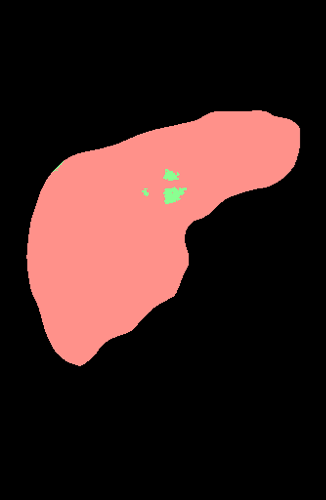}\label{fig:liver5}}\hfill
\subfigure{\fcolorbox{white}{red}{\includegraphics[width=0.132\textwidth]{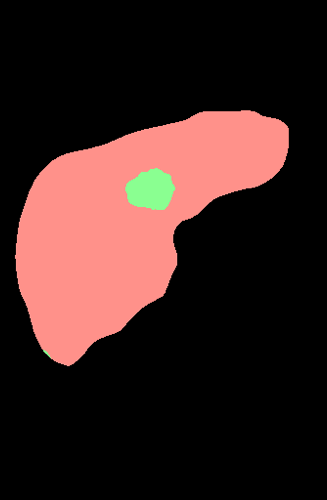}\label{fig:liver6}}}\hfill
\subfigure{\includegraphics[width=0.2\textwidth]{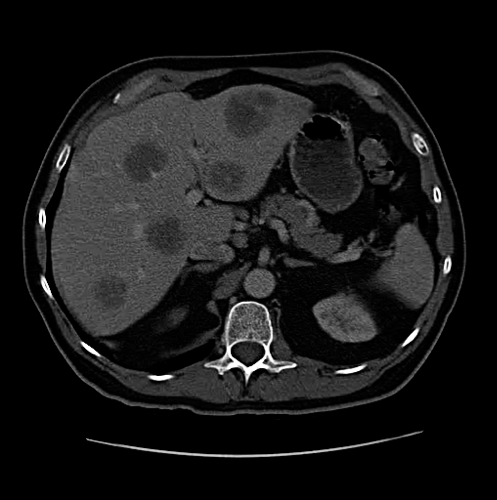}\label{fig:liver21}}\hfill
\subfigure{\includegraphics[width=0.2\textwidth]{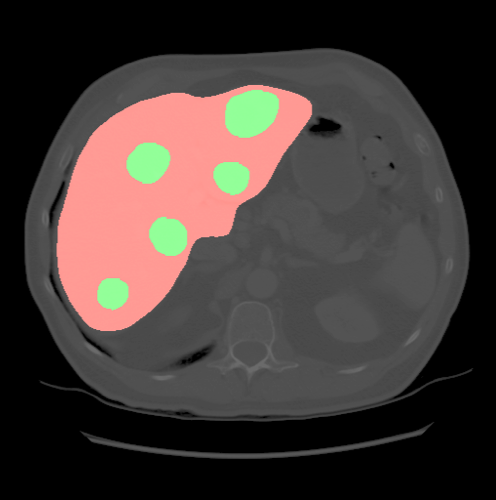}\label{fig:liver22}}\hfill
\subfigure{\includegraphics[width=0.132\textwidth]{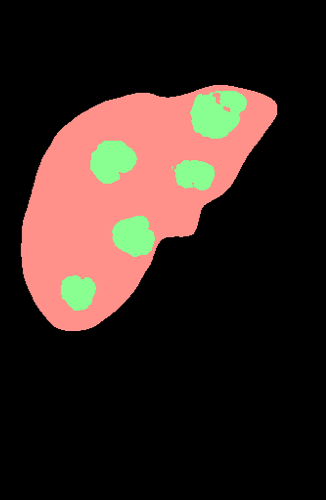}\label{fig:liver23}}\hfill
\subfigure{\includegraphics[width=0.132\textwidth]{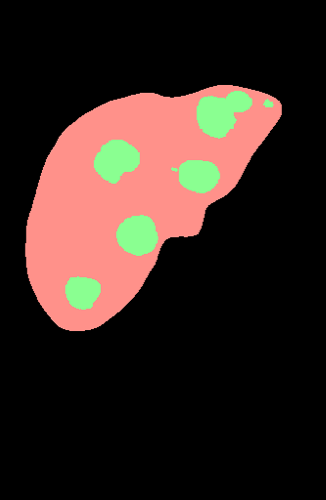}\label{fig:liver24}}\hfill
\subfigure{\includegraphics[width=0.132\textwidth]{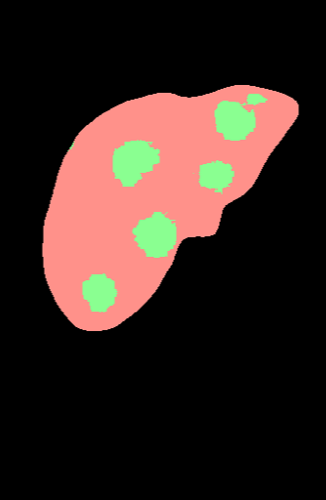}\label{fig:liver25}}\hfill
\subfigure{\fcolorbox{white}{red}{\includegraphics[width=0.132\textwidth]{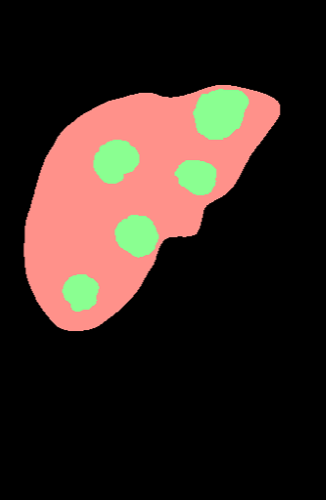}\label{fig:liver26}}}\hfill

\subfigure{\includegraphics[width=0.2\textwidth]{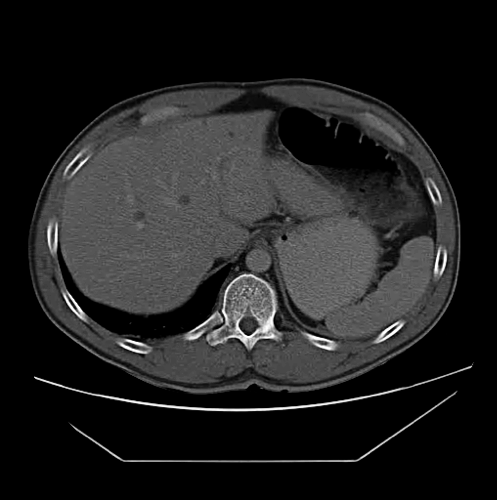}\label{fig:liver31}}\hfill
\subfigure{\includegraphics[width=0.2\textwidth]{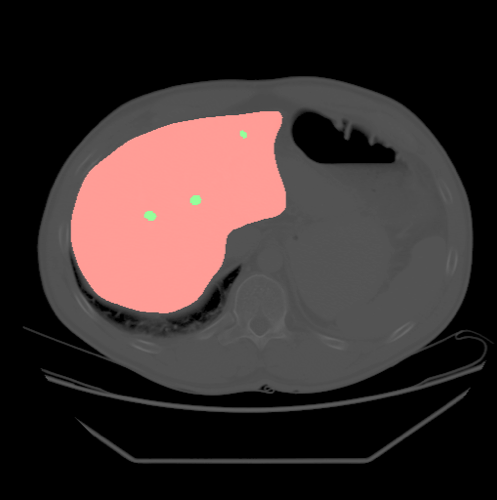}\label{fig:liver32}}\hfill
\subfigure{\includegraphics[width=0.132\textwidth]{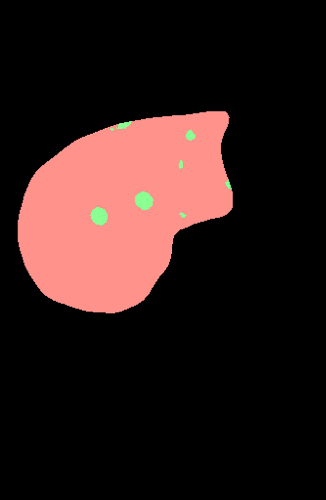}\label{fig:liver33}}\hfill
\subfigure{\includegraphics[width=0.132\textwidth]{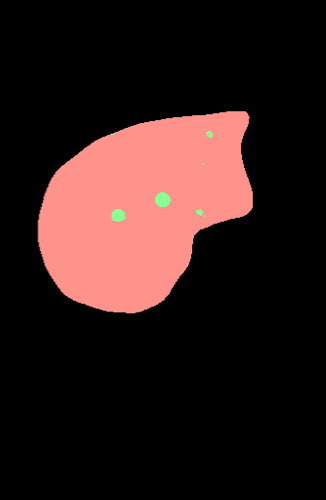}\label{fig:liver34}}\hfill
\subfigure{\includegraphics[width=0.132\textwidth]{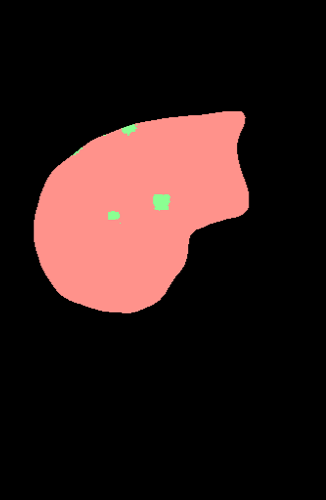}\label{fig:liver35}}\hfill
\subfigure{\fcolorbox{white}{red}{\includegraphics[width=0.132\textwidth]{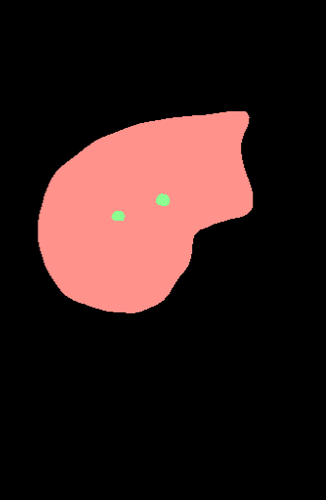}\label{fig:liver36}}}\hfill

\addtocounter{subfigure}{-18}
\subfigure[Input CT image]{\includegraphics[width=0.2\textwidth]{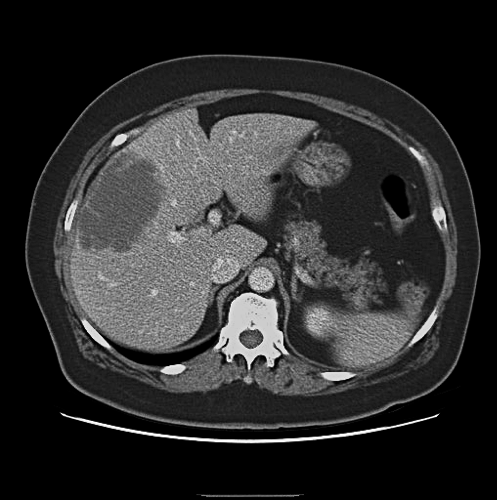}\label{fig:liver41}}\hfill
\subfigure[Ground Truth]{\includegraphics[width=0.2\textwidth]{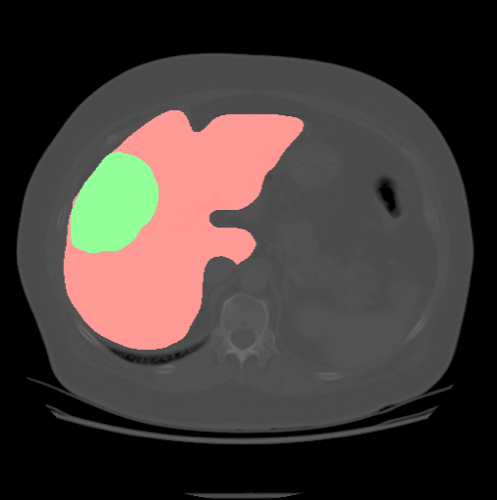}\label{fig:liver42}}\hfill
\subfigure[FCN8]{\includegraphics[width=0.132\textwidth]{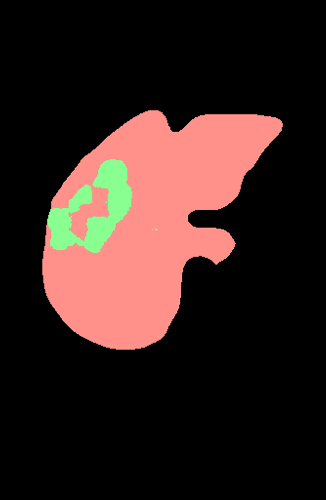}\label{fig:liver43}}\hfill
\subfigure[Unet]{\includegraphics[width=0.132\textwidth]{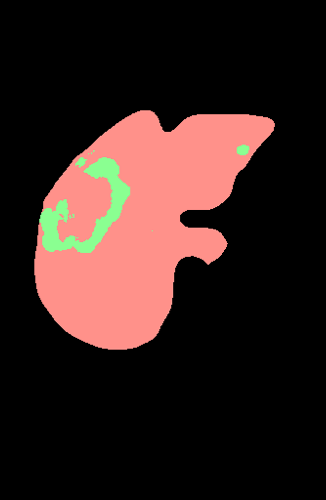}\label{fig:liver44}}\hfill
\subfigure[FC-ResNet]{\includegraphics[width=0.132\textwidth]{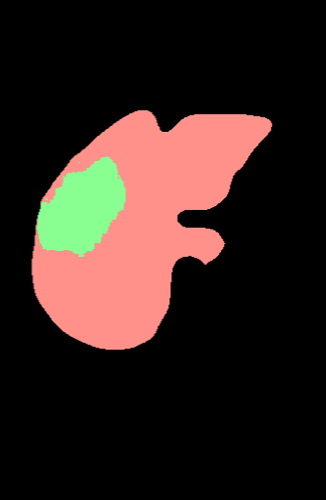}\label{fig:liver45}}\hfill
\subfigure[Ours]{\fcolorbox{white}{red}{\includegraphics[width=0.132\textwidth]{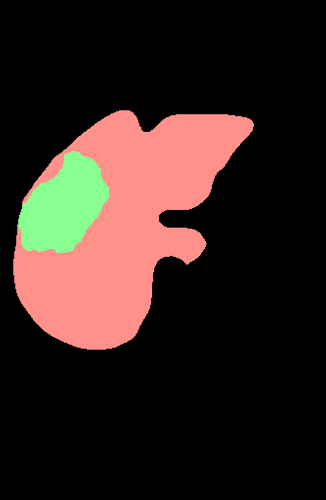}\label{fig:liver46}}}\hfill

\caption{Qualitative results on test set for the liver lesion dataset. Each line displays an example form the test set. From left to right: (a) represents an image, (b) displays the expert annotation of liver (red) and lesion (green), (c) displays a prediction for FCN8 model, (d) displays a prediction for UNet model, (e) displays a prediction for FC-ResNet model and (f) displays a prediction of our method.}
\label{fig:liver}
\vspace{-.5cm}
\end{figure*}

Finally, we show some qualitative results for liver lesion segmentation in Figure \ref{fig:liver}, obtained without any user interaction or manual intialization. Figure \ref{fig:liver41} displays sample input CT images to segment, Figure \ref{fig:liver42} shows ground truth annotations and Figures \ref{fig:liver43} to \ref{fig:liver46} present predictions of FCN8, UNet, FC-ResNet and of our approach respectively. Our approach performs better for all types of lesions: small lesions (see third row in Figure \ref{fig:liver}), medium size lesions (see first and second rows in Figure \ref{fig:liver}) and large lesions (see forth row in Figure \ref{fig:liver}). Furthermore, the lesion segmentation is better adjusted to ground truth annotation, has less false positives and does not have unsegmented holes or gaps within the lesions.

\subsection{Prostate dataset}
\label{sec:prostate_data}
The final experiment tested the segmentation framework on T2-w MR images of the prostate provided by the PROMISE12 challenge\footnote{https://grand-challenge.org/site/promise12/home/}. The training dataset contains 50 T2-w MR images of the prostate together with segmentation masks. The test set consists of 30 MR images for which the ground truth is held out by the organizer for independent evaluation. These MRIs are acquired in different hospitals, using different equipments, different acquisition protocols and include both patients with benign disease (e.g. benign prostatic hyperplasia) as well as with prostate cancer. Thus, the dataset variations include: voxel size, dynamic range, position, field of view and anatomic appearance. Contrary to all previously published methods we did not apply any pre-processing step nor volume resizing at training or testing time.

During training, we augmented the dataset using random sheering (with maximal range of $0.1$), rotations (with maximal range of $10$), random cropping ($256\times256$) and spline warping. We trained the model with RMSprop \cite{Tieleman2012} with an initial learning rate of $0.0004$, a learning rate decay of $0.001$ and a batch size of $24$. We used weight decay of $0.00001$. For each training, the model with the best validation Dice was stored. In total, we trained 10 models and averaged their outputs at the test time. Each time the model was trained, we randomly split $50$ training volumes into $40$ training images and $10$ validation images. Because the method is still based on 2D images, a connected component method was applied on the output to select the largest structure on each volume. 

\begin{table}[t!]
\begin{center}
\begin{tabular}{||c |c |c |c |c ||} 
\hline
\thead{Method} &
\specialcell{\thead{Score}\\{[-]}} &
\specialcell{\thead{Dice}\\{[\%]}} &
\specialcell{\thead{Avg. Dist.}\\{[mm]}} &
\specialcell{\thead{Vol. Diff.}\\{[\%]}}
\\
\hline
\multicolumn{5}{l}{\thead{2D FCNs}} \\
\hline
\textbf{Ours} & \textbf{83.02} & 87.4 & 2.17 & 12.37 \\%
\hline
SITUS & 79.92 & 84.13 & 2.96 & 23.00\\%
\hline
\multicolumn{5}{l}{\thead{3D FCNs}} \\
\hline
CUMED \cite{Yu17} & 86.65 & 89.43 & 1.95 & 6.95\\%
\hline
CAMP-TUM2 \cite{MilletariNA16} & 82.39 & 86.91 & 2.23 & 14.98 \\%
\hline
SRIBHME & 74.17 & 74.46 & 2.83 & 34.89\\%
\hline
\end{tabular}
\end{center}
\caption{Comparison to the automatic entries based on FCNs for the prostate dataset. For full ranking of all submitted methods please refer to the challenge website: \protect\url{https://grand-challenge.org/site/promise12/results/}.}
\label{tab:score_prostate}
\vspace{-.5cm}
\end{table}

Overall, for our method the Dice coefficient is of $87.4$ on the entire gland, with an average boundary distance of $2.17$mm and a volume difference of $12.37\%$. The comparison to other FCNs on the prostate data is shown in Table \ref{tab:score_prostate}. In comparison, we use the score provided by the challenge organizer. As it can be seen, our pipeline outperforms other method based on 2D FCNs and is competitive with methods based on 3D FCNs.

\begin{figure}[t!]
\centering
\subfigure[]{\includegraphics[width=0.16\textwidth]{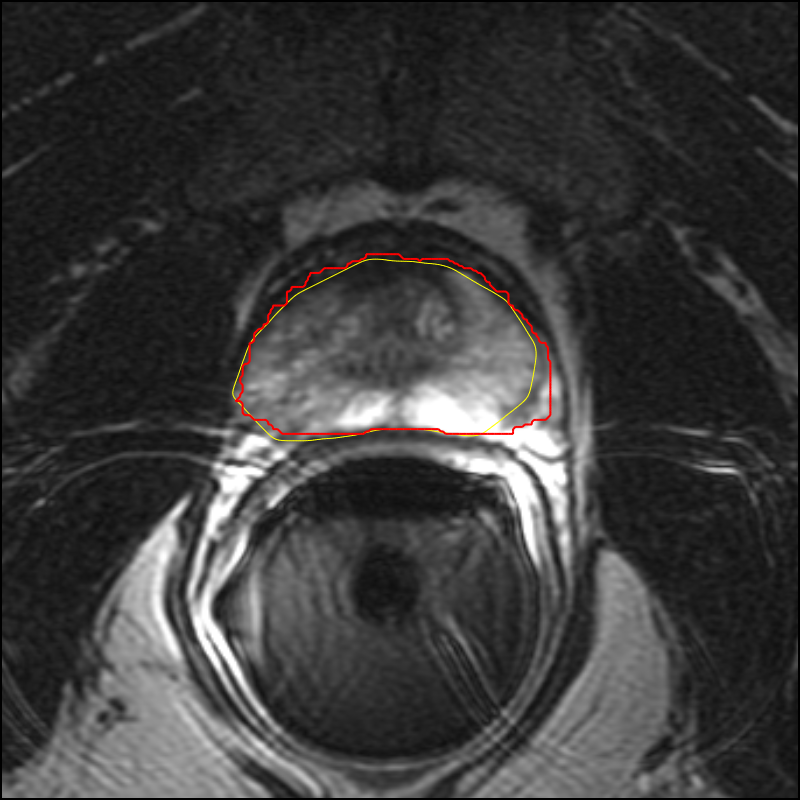}\label{fig:prostate_1}}\hfill
\subfigure[]{\includegraphics[width=0.16\textwidth]{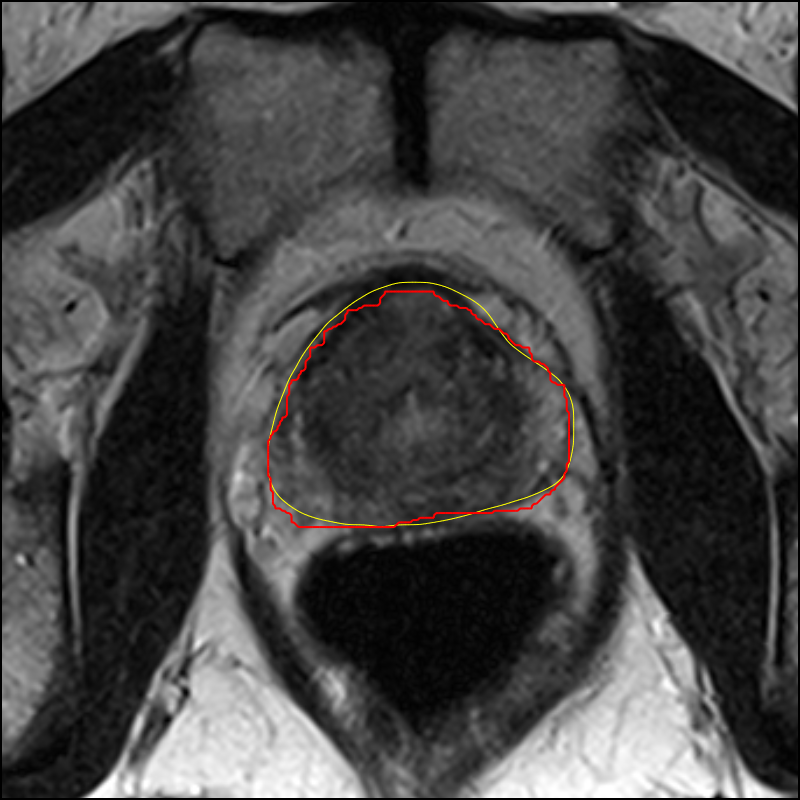}\label{fig:prostate_2}}\hfill
\subfigure[]{\includegraphics[width=0.16\textwidth]{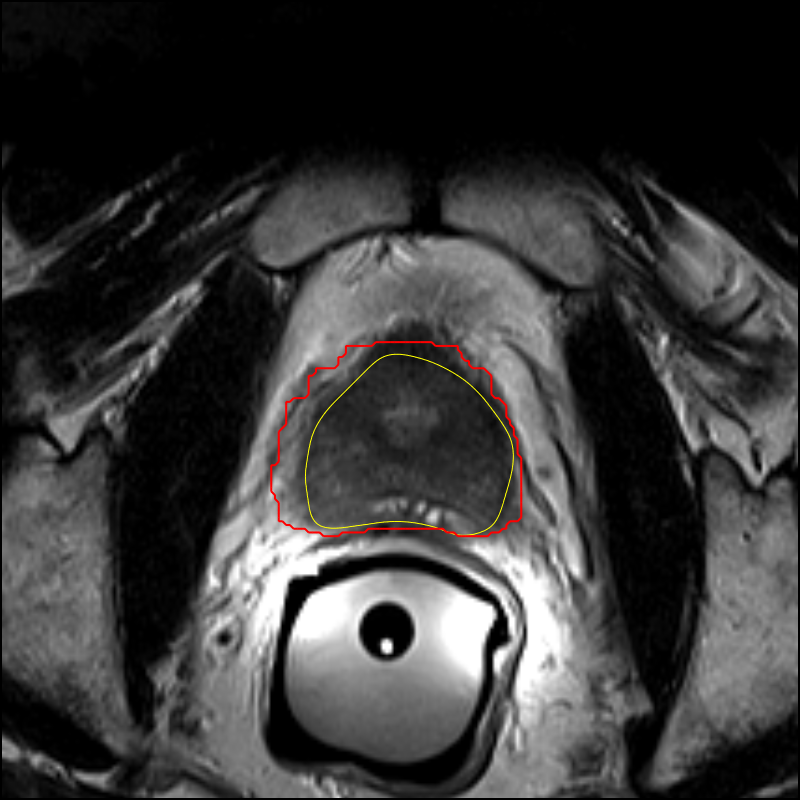}\label{fig:prostate_3}}\hfill
\caption{Qualitative results of our pipeline on the test set. The yellow line shows ground truth annotations while the red line displays our prediction. For more examples of qualitative results please refer to \protect\url{https://grand-challenge.org/site/promise12/resultpro/?id=UdeM2D&folder=20170201000157_2703_UdeM2D_Result}}
\label{fig:prostate}
\vspace{-.5cm}
\end{figure}

Figure \ref{fig:prostate} shows some qualitative results. The prostate segmentations are well adjusted to ground truth annotations and do not have unsegmented holes or gaps within the prostate. Due to the lack of 3D context information, our method, in some cases, is over-segmenting the base of the apex of the prostate (e.g. see Figure \ref{fig:prostate_3}). These results rank amongst the best automated approaches for prostate segmentation and without tedious and application specific pre-processing steps\footnote{For full ranking of all submitted methods, please refer to the challenge website: \protect\url{https://grand-challenge.org/site/promise12/results/}}.

\subsection{Data normalization}
\label{sec:normalizatoin}

\begin{figure*}[t!]
\centering
\subfigure[Prostate dataset]{\includegraphics[width=0.49\textwidth]{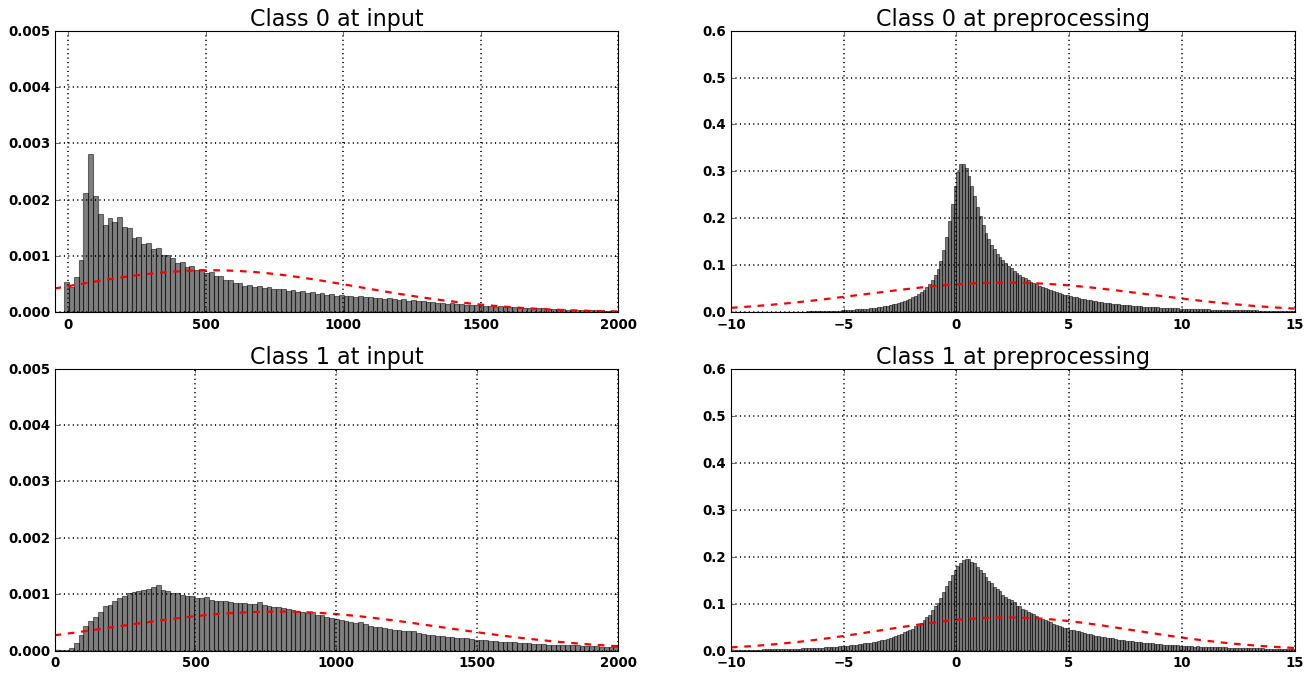}\label{fig:distribution1}}\hfill
\subfigure[Liver lesion dataset]{\includegraphics[width=0.49\textwidth]{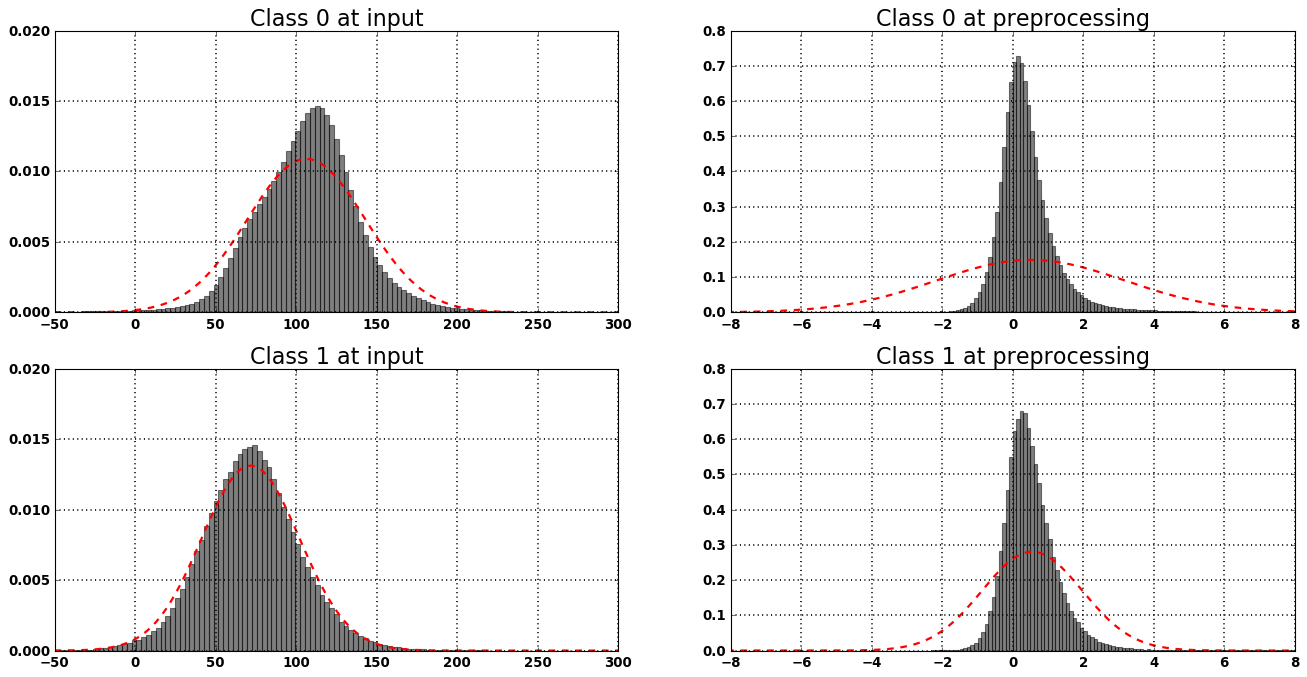}\label{fig:distribution2}}\hfill
\caption{Intensity distribution histograms for  (a) Prostate dataset and (b) liver lesion dataset. The plots represent the following information: (left-up subfigure) the 0-class distribution of input pixels, (left-bottom subfigure) the 1-class distribution for input pixels, (right-up subfigure) the 0-class distribution after FCN pre-processing, (right-bottom subfigure) the 1-class distribution after pre-processing. The red dashed-line represents normal distribution fitted to the data.}
\label{fig:distribution}
\vspace{-.5cm}
\end{figure*}

In this subsection, we provide a more detailed analysis on the trained models. In particular, we investigate the effect of an FCN-based pre-processor on the data distribution. The distribution of validation set pixel intensities at the input of our pipeline (input to FCN-based pre-processor) and at the input of the FC-ResNet are shown in Figure \ref{fig:distribution}. Together with intensity histograms, we plot the normal distribution fitting the validation data (dashed red line). 

Figure \ref{fig:distribution1} shows the plots for prostate data, where class 0 represent background and class 1 represent prostate. We observe that intensities are shifted from $[0, 2000]$ at the FCN input to $[-5, 10]$ at the FC-ResNet input. For liver lesion dataset, we only plot the distributions for liver (referred to as class 0) and lesion (referred to as class 1), ignoring the distribution of regions outside of the liver. The liver distributions are shown in Figure \ref{fig:distribution2}. We observe similar behavior as for the prostate data: intensities are shifted from $[0, 200]$ at the FCN input to $[-2, 3]$ at the FC-ResNet input.

\begin{figure*}[t!]
\centering
\subfigure[Input image]{\includegraphics[width=0.22\textwidth]{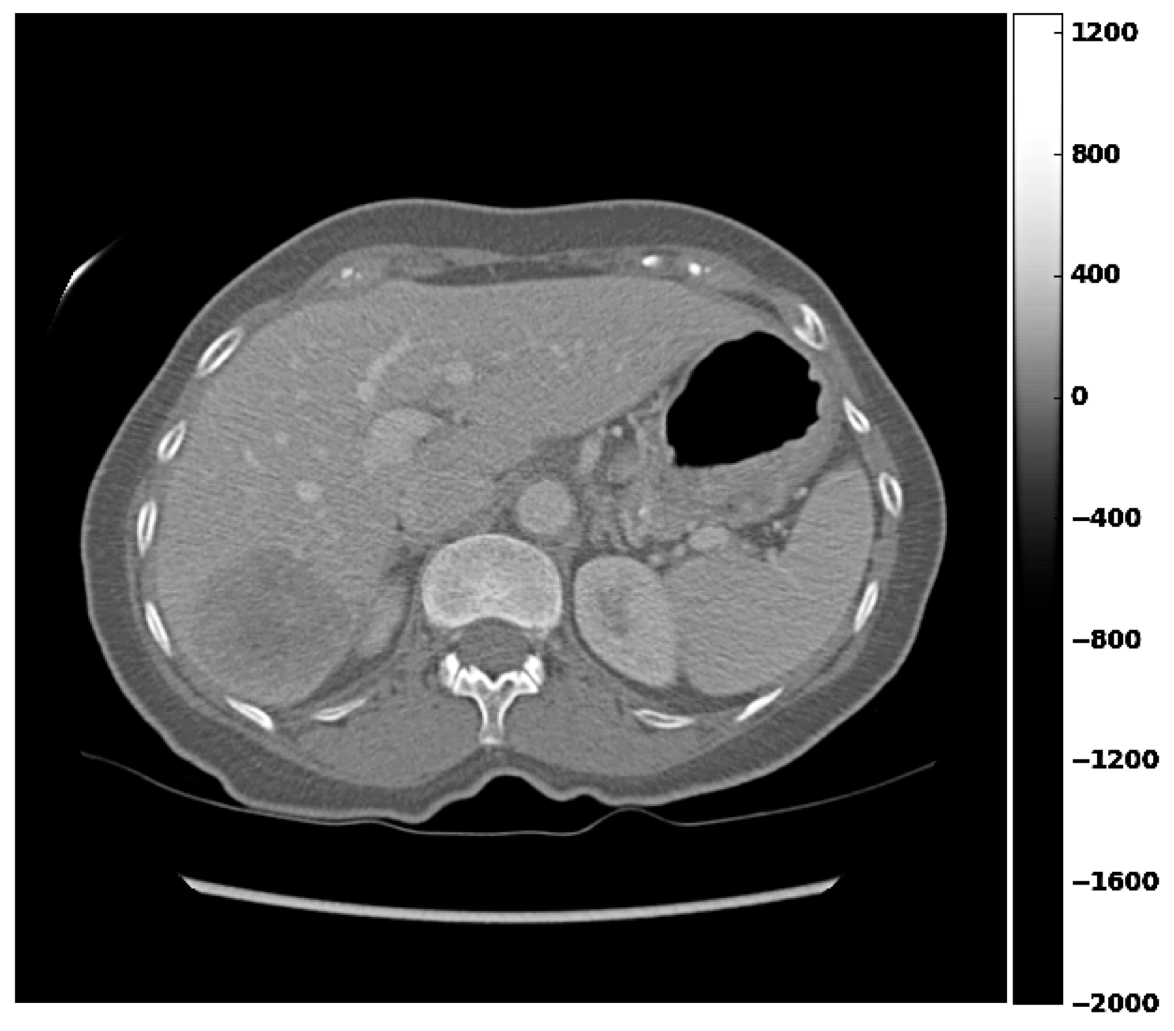}\label{fig:liver_preprocesor_vis_1}}\hfill
\subfigure[Pre-procesor output]{\includegraphics[width=0.22\textwidth]{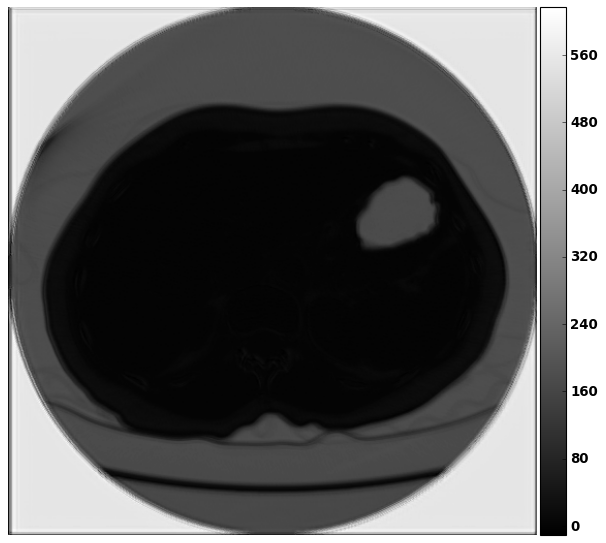}\label{fig:liver_preprocesor_vis_2}}\hfill
\subfigure[Image cropped to liver]{\includegraphics[width=0.22\textwidth]{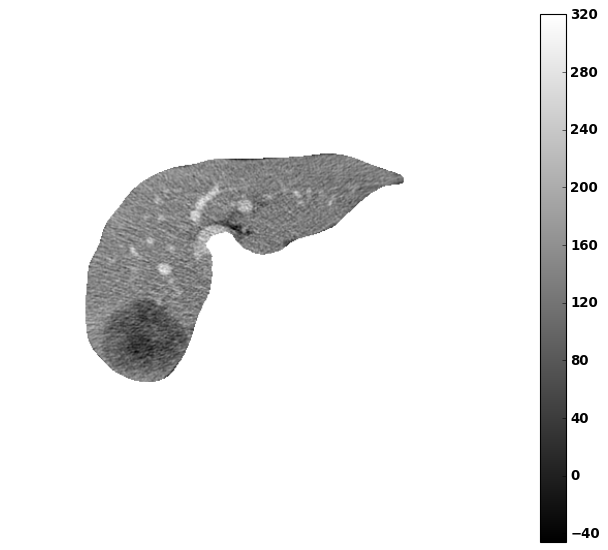}\label{fig:liver_preprocesor_vis_3}}\hfill
\subfigure[Pre-procesor output cropped to liver]{\includegraphics[width=0.22\textwidth]{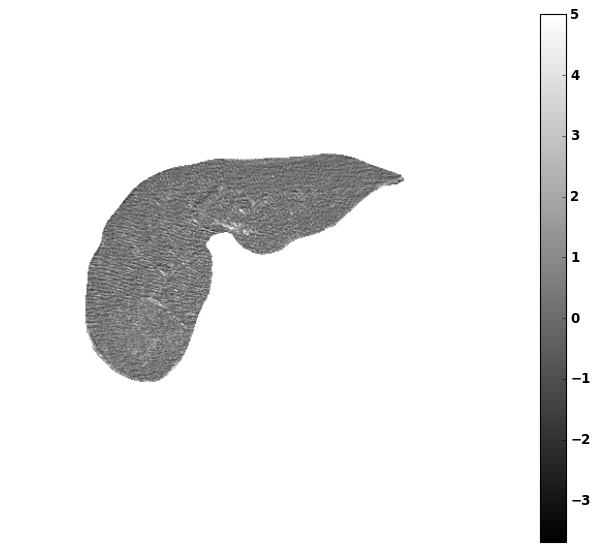}\label{fig:liver_preprocesor_vis_4}}\hfill

\caption{Visualization of the output of the pre-processor module for liver lesion dataset.}
\label{fig:liver_preprocesor_vis}
\end{figure*}

\begin{figure}[t!]
\centering
\subfigure[Input image]{\includegraphics[width=0.22\textwidth]{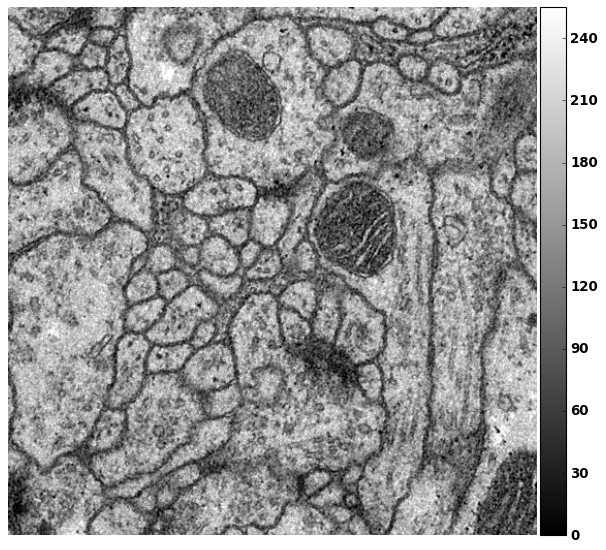}\label{fig:EM_preprocesor_vis_1}}\hfill
\subfigure[Pre-procesor output]{\includegraphics[width=0.225\textwidth]{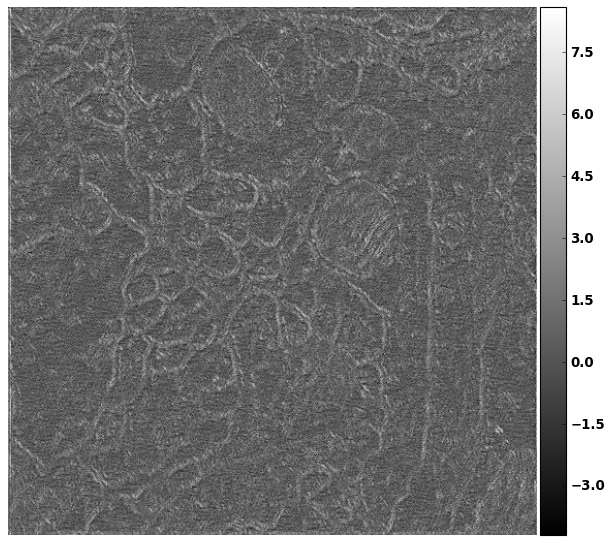}\label{fig:EM_preprocesor_vis_2}}\hfill
\caption{Visualization of the output of the pre-processor module for the EM dataset.}
\label{fig:EM_preprocesor_vis}
\vspace{-.5cm}
\end{figure}

The qualitative evaluation of the FCN-based pre-processor is displayed in Figures \ref{fig:liver_preprocesor_vis} and \ref{fig:EM_preprocesor_vis} for liver and EM data, respectively. Figure \ref{fig:liver_preprocesor_vis} shows visualizations of how a liver image is transformed from the input of our pre-processor (Figure \ref{fig:liver_preprocesor_vis_1}) to its output (Figure \ref{fig:liver_preprocesor_vis_2}). Analogously, Figures \ref{fig:liver_preprocesor_vis_3} and \ref{fig:liver_preprocesor_vis_4} emphasize how intensities within the liver change, by removing the void pixels of the image. Note that the FCN-based pre-processor does not perform any pre-segmentation of the lesion, i.e. the lesion is not more visible on the pre-processed image (Figure \ref{fig:liver_preprocesor_vis_3} than at the input (Figure \ref{fig:liver_preprocesor_vis_4})). This suggests that the FCN is only normalizing the data to values that are adequate for iterative refinement that happens within the FC-ResNet model. Similar observations can be made for EM data, which is displayed in Figure \ref{fig:EM_preprocesor_vis}. 

\section{Discussion \& Conclusion}
\label{sec:Conclusion}

In this paper, we have introduced a simple, yet powerful segmentation pipeline for medical images that combines fully convolutional networks with fully convolutional ResNets. Our pipeline is built from a low-capacity FCN model followed by a very deep FC-ResNet (more than 100 layers). We have highlighted the importance of pre-processing when using FC-ResNets and shown that a low-capacity FCN model can serve as a pre-processor to normalize raw medical image data. We argued that FC-ResNets are better complemented by the proposed FCN pre-processor than by traditional pre-processors, given the normalization they achieve. Finally, we have shown that using this pipeline we exhibit state-of-the-art performance on the challenging EM benchmark, improve segmentation results on our in-house liver lesion dataset, when compared to standard FCN methods and yield competitive results on challenging 3D MRI prostate segmentation task. These results illustrate the strong potential and versatility of the framework by achieving highly accurate results on multi-modality images from different anatomical regions and organs. 

The fact that FC-ResNets require some kind of data pre-processing is not very surprising, given the model construction. The identity path forwards the input data right to the output, while applying small transformations along the residual path. These small transformations are controlled by batch normalization. This input-forwarding especially affects medical imaging data, where the pixel intensities range is much broader than in standard RGB images. Therefore, adequate pre-processing becomes crucial for FC-ResNets to achieve significant improvement over standard FCN approaches, especially for some imaging modalities with greater variability in acquisition protocols. Standard FCNs do not contain the identity path and, thus, are more robust to the input data distribution. However, the lack of this identity path affects the optimization process and limits the depth of the tested models. 

In light of recent advancements in understanding both CNNs and ResNets, our pipeline can have two possible interpretations. On one hand, the pipeline could be explained as having the FCN model working as a pre-processor followed by the FC-ResNet model performing the task of an ensemble of relatively shallow models, leading to a robust classifier. On the other hand, the interpretation of our pipeline could revolve around the iterative inference point of view of ResNets \cite{Greff16}. In this scenario, the role of the FCN would be to produce an input proposal that would be iteratively refined by the FC-ResNet to generate the proper segmentation map. It is worth noting that, in both interpretations, FC-ResNets should be relatively deep (hundreds of layers) in order to take full advantage of the ensemble of shallow networks or iterative refinement of the initial proposal. FC-ResNets might not have been deep enough in many medical image segmentation pipelines to achieve these effects.

Potential future direction might involve experimentation with different variants of architectures that could serve as a pre-processor. This architecture exploration should not be limited to FCN-like models. From medical image segmentation perspective, the model could potentially benefit by expanding it to 3D FCN.

\section*{Acknowledgment}
\footnotesize
We would like to thank all the developers of Theano and Keras. We gratefully acknowledge NVIDIA for GPU donation. The authors would like to thank Mohammad Havaei and Nicolas Chapados for insightful discussions. The authors would like to thank Drs Simon Turcotte, R\'eal Lapointe, Franck Vandenbroucke-Menu and Ms. Louise Rousseau from the CHUM Colorectal, Hepato-Pancreato-Biliary Cancer Biobank and Database, supported by the Universit\'e de Montr\'eal Roger DesGroseillers Hepato-Pancreato-Biliary Surgical Oncology Research Chair, for enabling the selection of consenting patients with diagnostic imaging available for this study. This work was partially funded by Imagia Inc., MITACS (grant number IT05356) and MEDTEQ. An Tang was supported by a research scholarship from the Fonds de Recherche du Qu\'ebec en Sant\'e and Fondation de l'association des radiologistes du Qu\'ebec (FRQS-ARQ \#26993).

\ifCLASSOPTIONcaptionsoff
  \newpage
\fi



\bibliographystyle{ieee}
\bibliography{bib2}
%



\end{document}